\documentclass[runningheads]{llncs}
\usepackage{graphicx}
\usepackage{subcaption}
\begin{document}
\title{FlowChroma - A Deep Recurrent Neural Network for Video Colorization\thanks{Supported by Amazon Cloud Credits for Research.}}
%
\author{Thejan Wijesinghe\inst{1} \and
Chamath Abeysinghe\inst{1} \and
Chanuka Wijayakoon\inst{1} \and
Lahiru Jayathilake\inst{1} \and
Uthayasanker Thayasivam\inst{1}
}
\authorrunning{T. Wijesinghe et al.}
%
\institute{University of Moratuwa, Moratuwa, Sri Lanka
\email{\{chamath.14, thejanwijesinghe.14, chanuka.14, lahiruj.14, rtuthaya\}@cse.mrt.ac.lk}}
\maketitle              
\begin{abstract}
We develop an automated video colorization framework that minimizes the flickering of colors across frames. If we apply image colorization techniques to successive frames of a video, they treat each frame as a separate colorization task. Thus, they do not necessarily maintain the colors of a scene consistently across subsequent frames. The proposed solution includes a novel deep recurrent encoder-decoder architecture which is capable of maintaining temporal and contextual coherence between consecutive frames of a video. We use a high-level semantic feature extractor to automatically identify the context of a scenario including objects, with a custom fusion layer that combines the spatial and temporal features of a frame sequence. We demonstrate experimental results, qualitatively showing that recurrent neural networks can be successfully used to improve color consistency in video colorization.

\keywords{Video colorization  \and Image colorization \and Recurrent Neural Networks.}
\end{abstract}
\section{Introduction}
Colorizing a grayscale image to achieve a natural look has been a much-explored research problem in the recent years, especially with the rise of deep learning-based approaches for image processing. A primary goal has been to produce diverse colorizations, while also providing plausible colorizations that apply correct colors to identified objects. Desaturating an image is a surjective operation, but it is not injective. Hence, there are multiple possible colors to choose from when considering a pixel in a grayscale image - it is a one-to-many mapping.

Compared to the image colorization problem, colorizing black and white videos has largely been left behind. This problem has abundant training data, as one could easily convert a video to grayscale and test the colorization against the original video. Video colorization could be used as a video preprocessing technique, such as to enhance CCTV footage, and to restore old movies and documentaries. One could argue that video colorization could be taken as a direct extension of image colorization, where successive application of frame colorization would produce a colorized video. But obviously, there is no guarantee that the selected image colorization technique would color successive frames consistently, known as temporal coherence, since it would consider each frame as a separate task, ignoring the contextual connections between frames.  This would result in flickering colors, reducing the usefulness of such results.

The other prime obstacle has been the high computational costs in colorizing videos \cite{Levin:2004:CUO:1015706.1015780,6575125} - it adds another dimension across time on top of the already computationally intensive image colorization.

Furthermore, we observed that the most realistic image colorization results from current techniques are produced when some sort of human intervention is made, such as user scribbles that guide the colorization process \cite{DBLP:journals/corr/ZhangZIGLYE17,Levin:2004:CUO:1015706.1015780}. While this is feasible for a few images, it certainly does not scale up for videos with thousands of consecutive frames, as commercial videos run at 24 or more frames per second. Thus, efficiently colorizing a video with resource constraints and minimal supervision poses an interesting research problem.

There's a plethora of early video content shot in black and white that was enjoyed by older generations and remembered fondly. Such classical content is mostly forgotten and the later generations prefer colored content. Colorizing existing content is much cheaper than reproducing them entirely in color today. 

Our research contributions are as follows;
\begin{enumerate}
    \item We propose a new fully automated video colorization framework focusing on improved temporal and contextual coherence between frames and scene changes. 
    \item We use a Recurrent Neural Network (RNN) based architecture to maintain contextual information across frames for consistent coloring.
    \item We study the effects of using RNNs on the colorization of videos.
\end{enumerate}

\section{Related Work}

Most of the previous work in the colorization domain has been done for image colorization, and video colorization is now gaining momentum with their success. The current image colorization algorithms can broadly be put into two major categories: parametric methods \cite{DBLP:journals/corr/abs-1712-03400,DBLP:journals/corr/ChengYS16,7410429,Iizuka:2016:LCJ:2897824.2925974,DBLP:journals/corr/ZhangIE16,DBLP:journals/corr/ZhangZIGLYE17,DBLP:journals/corr/LarssonMS16} and non-parametric methods \cite{10.1007/978-3-540-88690-7_10,Chia:2011:SCI:2070781.2024190,Gupta:2012:ICU:2393347.2393402,Huang:2005:AED:1101149.1101223,Irony:2005:CE:2383654.2383683,Levin:2004:CUO:1015706.1015780,1467343,Welsh:2002:TCG:566654.566576,1621234,Luan:2007:NIC:2383847.2383887,Morimoto:2009:ACG:1599301.1599333,qu-2006-manga}. Parametric methods learn predictive functions from large datasets of color images; once the predictive function's parameters are learned with an appropriate optimization objective, it is ready to predict colors in a fully automatic manner. Alternatively, non-parametric methods require some level of human intervention.  

There are mainly two non-parametric methods explored in the literature: scribble-based and transfer-based. Scribble-based colorization schemas \cite{Huang:2005:AED:1101149.1101223,Levin:2004:CUO:1015706.1015780,1621234,Luan:2007:NIC:2383847.2383887,qu-2006-manga} require manually chosen color scribbles on the target grayscale image. In few instances, scribble-based colorization methods are extended to video colorization as well \cite{Levin:2004:CUO:1015706.1015780,1621234}. Transfer-based colorization schemas \cite{10.1007/978-3-540-88690-7_10,Chia:2011:SCI:2070781.2024190,Gupta:2012:ICU:2393347.2393402,Irony:2005:CE:2383654.2383683,1467343,Welsh:2002:TCG:566654.566576,Morimoto:2009:ACG:1599301.1599333} require the user to select semantically similar colorful reference images to match similar segments of the target grayscale image. 

Applying non-parametric methods on both image and video colorization has a number of drawbacks, the most prominent among which is the inability to fully automate the colorization process. In color transferring approaches, there is a manual intervention in searching for colorful reference images. Scribble-based colorization may require tens of well-placed scribbles plus a carefully chosen, rich pallet of colors in order to achieve convincing, natural results for a complex image.  

Both scribble-based and transfer-based video colorization schemas can only be automated within a frame sequence without a scene change; i.e. at each scene change, if the process is scribble-based, the user will have to introduce a new set of scribbles. If it is transfer-based, a new selection of swatches with or without a new reference image will be required. 

Comparatively, parametric colorization schemas can fully automate the colorization process. Deshpande et al. \cite{7410429} proposed a parametric image colorization schema which formulates the colorization problem as a quadratic objective function and trained it using the LEARCH framework \cite{Ratliff2009}. With the unparalleled success of deep neural networks, solutions that utilize DNNs have been proposed as parametric image colorization schemas. Cheng et al. \cite{DBLP:journals/corr/ChengYS16} proposed an image colorization schema which leverages a three-layer fully connected neural network that was trained by the inputs of a set of image descriptors: luminance, DAISY features \cite{4587673} and semantic features.  More recently,  many authors have employed convolutional neural netowrks(CNN) and generative adversarial networks (GAN) in their colorization schemas rather than conventional deep neural networks (DNN). Zhang et al. \cite{DBLP:journals/corr/ZhangIE16} proposed a CNN-based colorization schema which predicts a probability distribution of possible colors for each pixel in order to address the typical ambiguous and multimodal nature of image colorization \cite{10.1007/978-3-540-88690-7_10}.

They also  introduced a CNN based color recommender system \cite{DBLP:journals/corr/ZhangZIGLYE17} that propagates user-provided scribbles while satisfying high level color preferences of the user. Larsson et al. \cite{DBLP:journals/corr/LarssonMS16} trained an end-to-end network to predict colors of an image with the hypercolumns \cite{DBLP:journals/corr/HariharanAGM14a} for each pixel generated from a pre-trained VGG-16 network without a classification layer. Iizuka et al. \cite{Iizuka:2016:LCJ:2897824.2925974} proposed a colorization method that utilizes a CNN based architecture, combining a high-level semantic feature extractor, a mid-level feature network and a colorization network. More recently, inspired by the colorization model of Iizuka et al. \cite{Iizuka:2016:LCJ:2897824.2925974}, Baldassarre et al. \cite{DBLP:journals/corr/abs-1712-03400} replaced the high-level semantic feature extractor in the colorization model of Iizuka et al. \cite{Iizuka:2016:LCJ:2897824.2925974} with a pre-trained CNN image classifier: Inception-ResNet-v2 \cite{DBLP:journals/corr/SzegedyVISW15}. This transfer learning approach significantly reduces the computational time as well as the need for extreme amounts of data and hardware resources to train the colorization network to yield a quality colorization result.

Most of the fully-automatic, parametric image colorization solutions can be extended to video colorization domain by treating a video merely as a sequence of independent frames. But considering video frames independently causes colors to shift erratically, failing to maintain temporal coherence throughout the frame sequence, causing visual fatigue for viewers. For an example, a wall in one frame may be colored in one shade of yellow and the same wall should maintain that color in subsequent frames, rather than changing to a shade of white. Failing to capture these details drastically reduces the quality of colored videos, because the user can notice color mismatches and flickering between video frames. In this research, we explore the effectiveness of employing RNNs to preserve the temporal coherence in video colorization while mitigating the challenges of computational time and need for large amounts of data, with the help of a transfer learning application.

\section{Proposed Approach}
When modeling the video colorization problem as a learnable function, we have chosen the CIE La*b* color space to represent video frames. According to Ruderman et al.\cite{Ruderman98statisticsof}, La*b* color space was developed to minimize correlation between the three coordinate axes of the color space. La*b* color space provides three decorrelated, principal channels corresponding to an achromatic luminance channel L and two chromatic channels as a* and b*. 
If we have a grayscale frame, that means we already have the luminance layer of that particular frame, the next step is finding a plausible a*, b* combination and fusing them together to come up with a final colored frame, given that there is temporal coherence when predicting a* and b* combinations. Therefore, the main assumption here is that for every luminance component of video frames \begin{equation} X\textsubscript{t}^L\in R\textsuperscript{H$\times$W$\times$1} \end{equation} there exists a function F such that \begin{equation}F: \{X\textsubscript{t}^L, X\textsubscript{t-1}^L,..., X\textsubscript{t-(T-1)}^L\} \rightarrow (X\textsubscript{t}\textsuperscript{a$^\ast$}, 
X\textsubscript{t}\textsuperscript{b$^\ast$})\end{equation}

Here, $X_{t}\textsuperscript{k}$ represents the $a$ or $b$ color layer in $t\textsuperscript{th}$ time frame, while H, W and T represent frame height, width and total number of previous frames used for prediction, respectively.

The chromatic channels a* and b* define an Euclidean space where the distance to the origin determines the chroma. Change of values in one channel imposes minimal effect on values of the other two. This decorrelation of the three channels allows us to combine the luminance with the predicted chromatic channels, ensuring an image construction with high level of detail but with almost non-existent cross-channel artifacts.

\subsection{Proposed Architecture}

\begin{figure}[!h]
	\includegraphics[width=\textwidth]{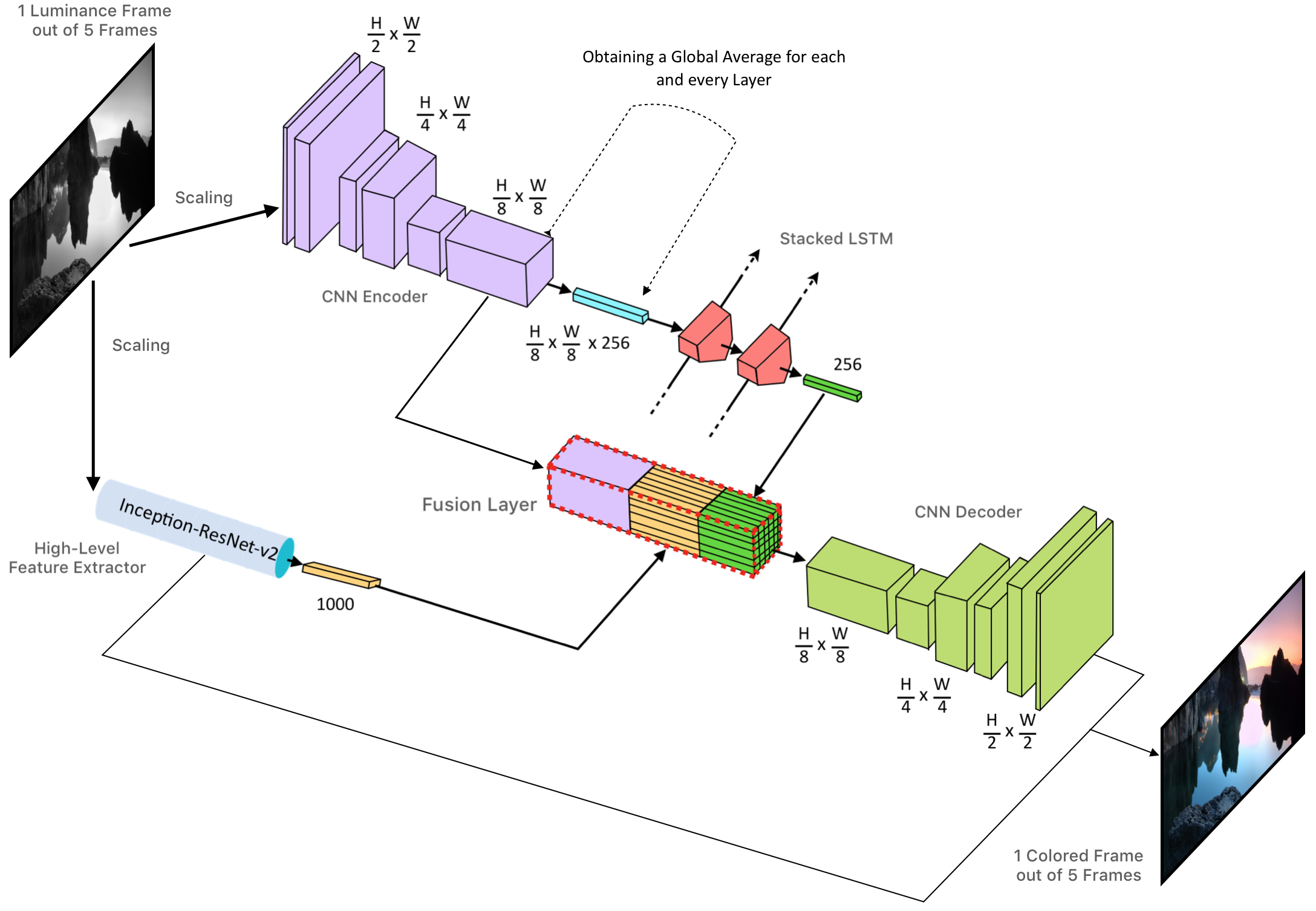}
	\caption{FlowChroma Architecture Diagram}
	\label{fig:flowchroma-architecture}
\end{figure}

FlowChroma architecture can be divided into five main components, as shown in Figure \ref{fig:flowchroma-architecture}: the CNN encoder, global feature extractor, stacked LSTM, fusion layer and the CNN decoder. We include Inception-ResNet-v2 network as a global feature extractor; this is a transfer learning technique, drawing inspiration from the works of Iizuka et al. \cite{Iizuka:2016:LCJ:2897824.2925974} and Baldassarre et al. \cite{DBLP:journals/corr/abs-1712-03400} This significantly reduces the computational complexity in training the model. 

Although the use of Long Short-Term Memory (LSTM) units \cite{10.1162/neco.1997.9.8.1735} to support video colorization has been proposed before, this is one of the first architectures to produce experimental results showing the effectiveness of it specifically towards video colorization. An LSTM is a special form of recurrent neural networks (RNNs). All RNNs have loops within their architecture, acting as a memory cell allowing information to persist for a certain period. They are able to connect previously learned information to the present task. LSTMs specifically outperform regular RNNs in many scenarios, as they have a superior ability to learn longer-term dependencies against vanilla RNNs. When considering an arbitrary frame sequence, it can include scene changes as well. Therefore, our model also needs to learn how much it should remember or forget while generating a frame sequence - this criteria makes LSTMs an ideal candidate for our use case over vanilla RNNs.

As shown in Figure \ref{fig:flowchroma-architecture}, the CNN encoder extracts local features such as texture and shapes while the Inception-ResNet-v2 extracts high level semantic information such as objects and environments from an individual frame. A stacked LSTM is being employed to grasp temporal features of a sequence of frames. The outputs from the CNN encoder, Inception network and the LSTM are then fused together in the fusion layer to provide inputs to the colorization network or the CNN decoder. The CNN decoder is used to predict a* and b* layers related to the input luminance frame at the current time step in a spatio-temporal manner.

\begin{figure*}[htb]
	\includegraphics[width=\textwidth]{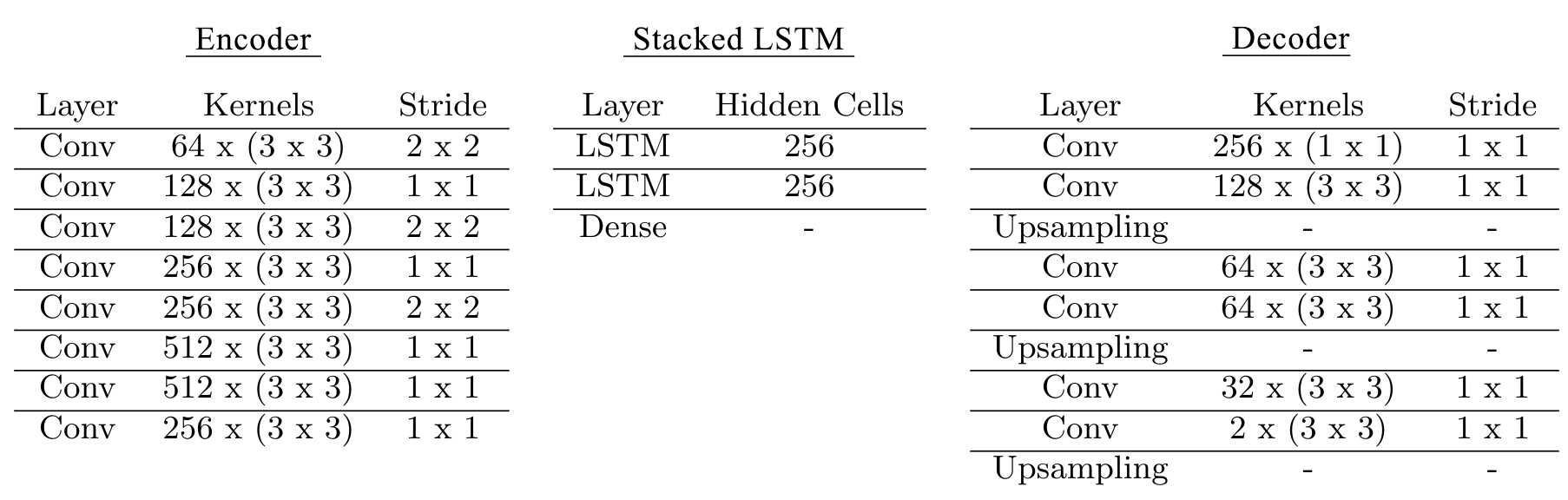}
    \caption{FlowChroma Architecture: The CNN encoder extracts local features while the Inception network extracts high level semantic information from a frame. The stacked LSTM grasps temporal features from a sequence of frames. The outputs from the CNN encoder, Inception network and the LSTM are then fused together in the fusion layer to provide inputs to the colorization network or the CNN decoder. Note that the CNN encoder, decoder, fusion layer and Inception network are all applied to every temporal slice of the input.}
\end{figure*}

\subsection{Grasping Local \& Global Features of each Individual Frame}
In order to grasp local features such as shapes in frame at each time step, we apply a CNN encoder to every temporal slice of the input. It processes a $t\times H\times W$ grayscale frame sequence and outputs a sequence of $t\times H/8\times W/8\times 256$ feature encodings.

Global features such as objects and environments are helpful for the CNN decoder to provide an appropriate colorization. The high-level feature extractor is a pre-trained Inception-Resnet-v2 model without the last SoftMax layer. When training FlowChroma, we keep Inception's weights static. At each time step, we scale the input luminance frame to $299\times 299$, and then stack itself to obtain a three channel frame in order to satisfy Inception's input dimensionality requirements. Then we feed the resultant frame to Inception and obtain its logits output (the output before the softmax layer). When the results at each time step are combined, we get a final embedding of $t\times 1000$ for the entire sequence.

\subsection{Capturing Temporal Features}
In order to grasp temporal variations of the frame sequence, we use a 2-layer stacked LSTM model. The CNN encoder provides a local feature encoding of  $t\times H/8\times W/8\times 256$. By employing global average pooling operation on that encoding at each time step, we obtain an embedding of $t\times 256$, which can be used as inputs to the stacked LSTM. Stacked LSTM has two LSTM layers, each having 256 hidden states, thus giving us an output with the dimensions of $t\times 256$. This output improves temporal coherence of the video colorization predictions.

\subsection{Fusing Local and Global Spatial Features with Temporal Features}
\begin{figure}[h]
  \centering
  \includegraphics[width=0.5\textwidth]{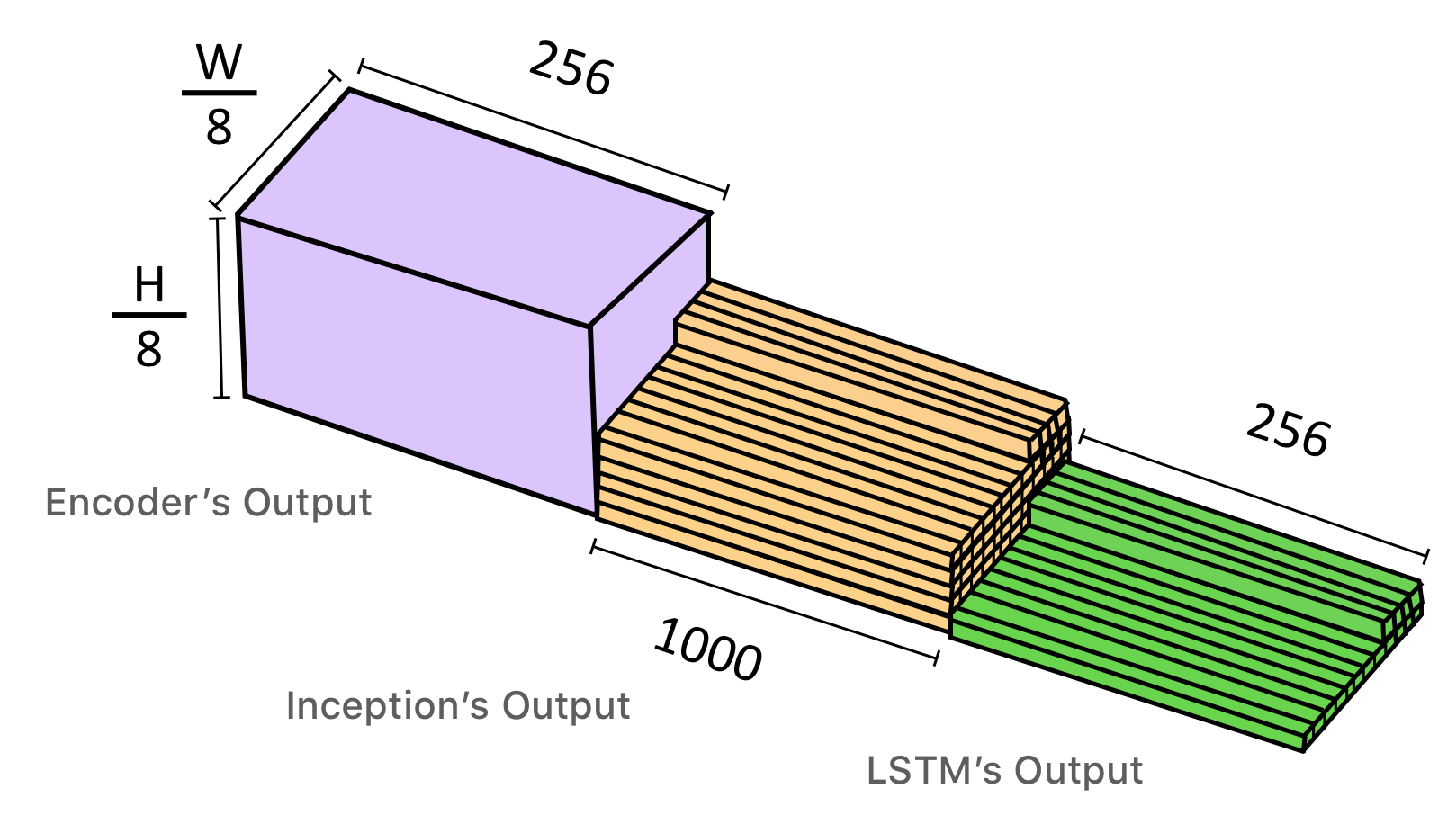}
  \caption{Fusion Layer - the outputs of the Inception network and the LSTM are replicated and stacked with the CNN encoder's output.}
  \label{fig:fusion-layer}
\end{figure}

Fusing local and global level spatial features with temporal features will be done by a specially crafted fusion layer, first introduced by Iizuka et al. \cite{Iizuka:2016:LCJ:2897824.2925974} Similar to CNN encoder, we apply the fusion layer to every temporal slice of the input. The fusion layer takes the output embeddings from Inception and stacked LSTM to replicate it $H/8\times W/8$ times and then concatenates them with the output provided by the CNN encoder. The fusion mechanism is more comprehensively illustrated in Figure \ref{fig:fusion-layer}.

\subsection{Colorization Decoder Network}
Once the local and global spatial features are fused with temporal features, they are processed by a set of convolutions and up-sampling layers in the CNN decoder. Similar to the CNN encoder and Fusion layer, we apply the CNN decoder to every temporal slice of the input. The decoder takes a $t\times H/8\times W/8\times 1512$ input and results in a final output with dimension of $t\times H\times W\times 2$. The resultant sequence can be considered as the sequence of a* and b* layers for the input sequence of luminance frames, once this result is appropriately merged with the input sequence, we can obtain the final colorized frame sequence.

\subsection{Optimization and Learning}
Optimal model parameters were found by minimizing an objective function defined over predicted outputs and actual results. To quantify the loss, mean squared error between each pixel in a*, b* layers of predicted and actual results were used. If we consider a video V, the MSE loss is estimated by,

\begin{equation}
\small
C(X,\Theta ) = \frac{1}{2nHW}\sum_{t=0}^{n}\sum_{k\in {a,b}}\sum_{i=1}^{H}\sum_{j=1}^{W} (X\textsuperscript{k}_{t_{i,j}} - \hat{X}\textsuperscript{k}_{t_{i,j}}) ^ 2
\end{equation}

Here $\theta$ represents all model parameters and $X\textsuperscript{k}_{t_{i,j}}$ represents the $(i,j)$ pixel in $t\textsuperscript{th}$ time frame's $k$ layer. This objective function can be extended to batch level by taking the average. 

\begin{equation}
C(X,\beta) = \frac{1}{\left | \beta  \right |}\sum_{X\in {\beta}} C(X,\Theta)
\end{equation}
To optimize the above objective function, we used Adam optimizer\cite{DBLP:journals/corr/KingmaB14}.

\subsection{Training}

FlowChroma was trained for roughly 50 hours on 50,000 short, preprocessed video clips, taken from the FCVID \cite{FCVID} video dataset. Videos were randomly selected from the dataset, converted to LAB color space and resized to match the input shape of the network.  We used a batch size of 20 and a validation split of 10\%. Training was done on an AWS EC2 instance that had 32 virtual CPUs and four NVIDIA Tesla P100 GPUs, with a total video memory of 64 GB.

\section{Experiments}
We compare FlowChroma's video colorization performance by taking the Deep Koalarization framework proposed by Baldassarre et al. \cite{DBLP:journals/corr/abs-1712-03400} as our baseline model. There are mainly two reasons for this choice, rather than another image colorization framework or a state-of-the-art technique.
\begin{enumerate}
\item Both FlowChroma and Deep Koalarization use the same transfer learning application of obtaining global features of an image or a video frame from a pre-trained object classifier and fusing them in the fusion layer, similar to Iizuka et al. \cite{Iizuka:2016:LCJ:2897824.2925974}
\item The main purpose of our research is emphasizing the use of sequence models in preserving temporal coherence between frames and scene changes rather than extremely realistic colorizations; to achieve that, comparison of our framework with a good enough image colorization framework is sufficient.
\end{enumerate}

To evaluate the performance of FlowChroma against our baseline model, we randomly selected 1,000 videos from the FCVID dataset, belonging to various categories depicting a wide range of scenarios, derived their grayscale variants and colorized them with the two models.

In order to provide a fair comparison of the two model's colorization performance, we used Inception-ResNet-v2, pre-trained object classifier as the global feature extractor for both FlowChroma and the baseline model. We also trained both models on the same dataset and hardware environment upto a close validation loss. Subsequently, a qualitative assessment of the colorizations was performed. 

Our model only takes a sequence of 5 frames as an input at once, but when running inference we need to colorize videos with hundreds of frames. Thus, we use a sliding window approach during inference. In contrast to that, our baseline model only takes a single frame as input at a time, thereby coloring each frame in a video independently.

We first confirm that our model performs well in colorization, and verify that although we use a recurrent architecture, it still converges. Next, we show that we can achieve temporal and contextual coherence through video frames with LSTMs. Finally, we discuss the weaknesses of the proposed architecture and discuss possible solutions.

\section{Results and Discussion}

\begin{figure}[!h]
  \centering
  \subcaptionbox{\label{fig:parachute}}
    {\includegraphics[width=0.45\textwidth]{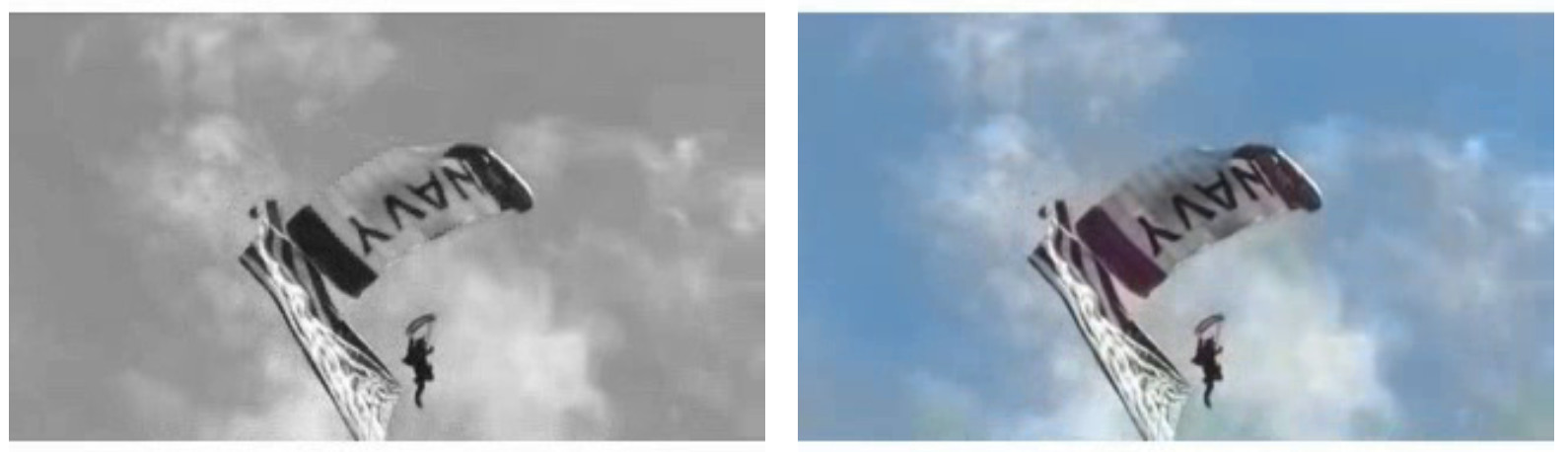}}
  \subcaptionbox{\label{fig:fifa}}
    {\includegraphics[width=0.45\textwidth]{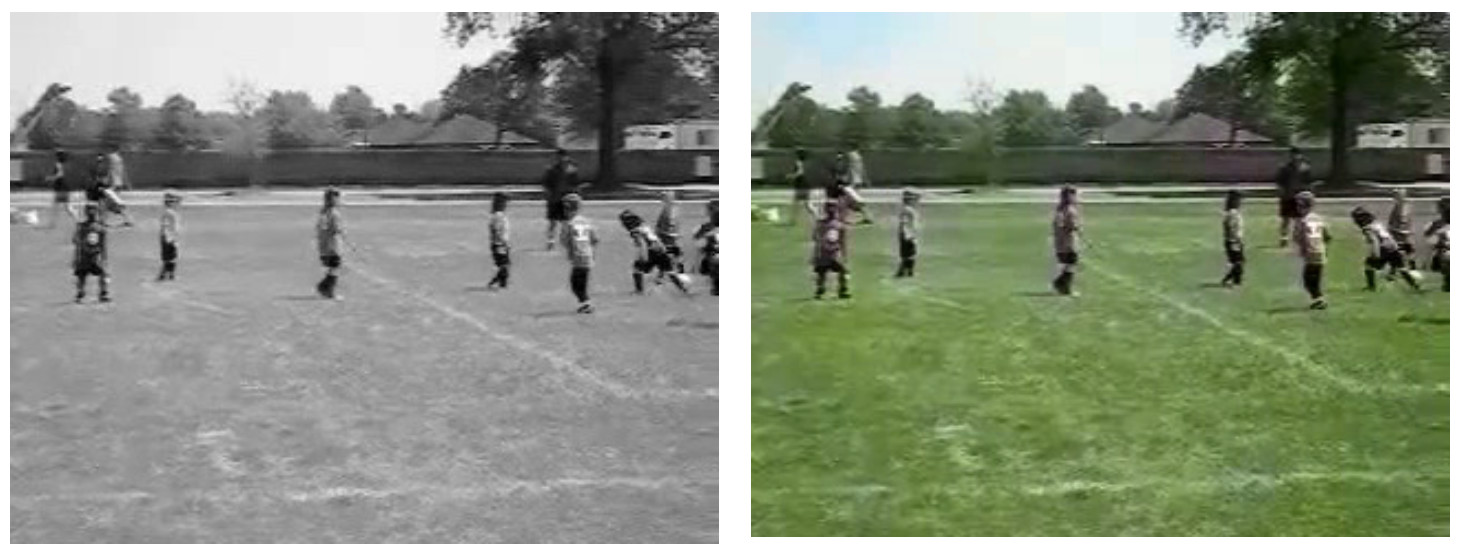}}
  \subcaptionbox{\label{fig:lady-face}}
    {\includegraphics[width=0.5\textwidth]{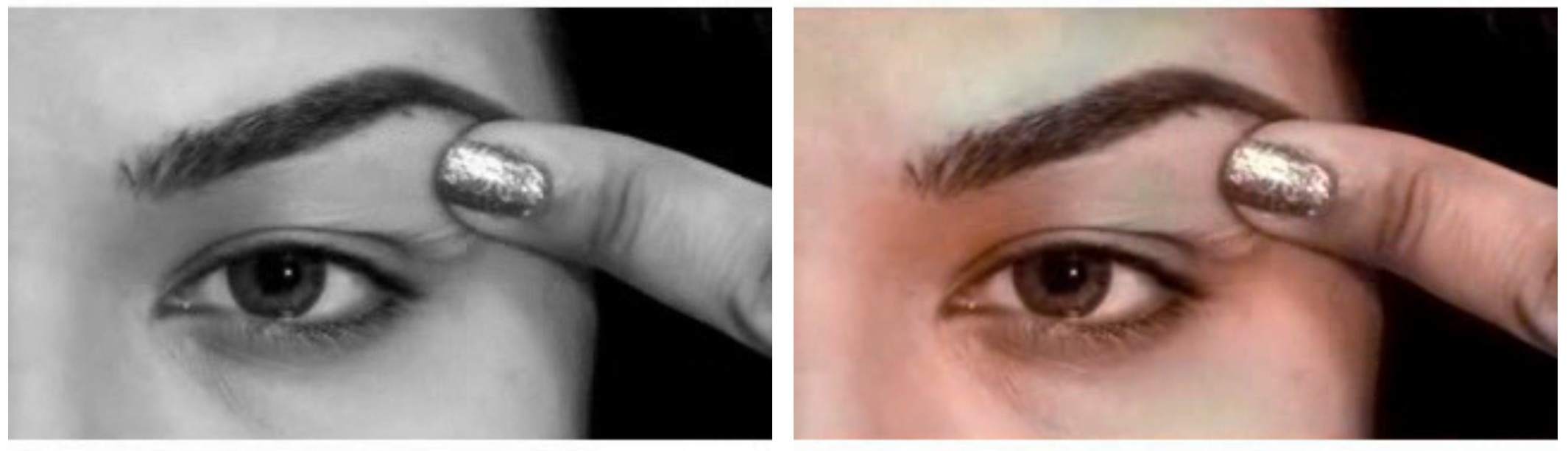}}
  \caption{FlowChroma generalizes commonly encountered scenes and objects and assigns them appropriate colors during inference. It also generates an acceptable variation of colors in each scene throughout the colorization results, as demonstrated in \ref{fig:parachute}, \ref{fig:fifa} and \ref{fig:lady-face}. In \ref{fig:parachute}, note how the parachute and the flag are colored in red hues while the sky is in blue. In \ref{fig:lady-face}, the eye color and skin tones over different regions in the face make the frame appear more realistic.}
\end{figure}

In general, we observed that our model produces appropriate colorization results, assigning realistic colors to objects within each scenario. Furthermore, the system successfully maintains color information between frames, keeping a natural flow and a high spatio-temporal coherence at both global and local levels for videos with common objects and environments in the training dataset. We also observed that for sudden or quick object movements, our model added blobs of flickering that followed the movement of the object.

In terms of raw colorization, our model generalizes commonly encountered scenes and objects and assigns them appropriate colors during inference. Figure \ref{fig:fifa} depicts a scene with a large field in the foreground and the sky in the background. This type of colorizations are observed throughout the results, and stand to reason that the system generalizes the scenes in the training dataset.

We observe LSTM's sequence learning capabilities on colorization at two scales; locally and globally. At a global scale, FlowChroma maintains the overall color composition of a scene throughout the video better than the baseline image colorization model. At a local level, the baseline model sometimes mistakenly colorizes small regions of a frame with inappropriate colors, but FlowChroma avoids such mistakes.

\begin{figure*}[!h]
    \centering
    \subcaptionbox{\label{fig:elephants}}
	  {\includegraphics[width=\textwidth]{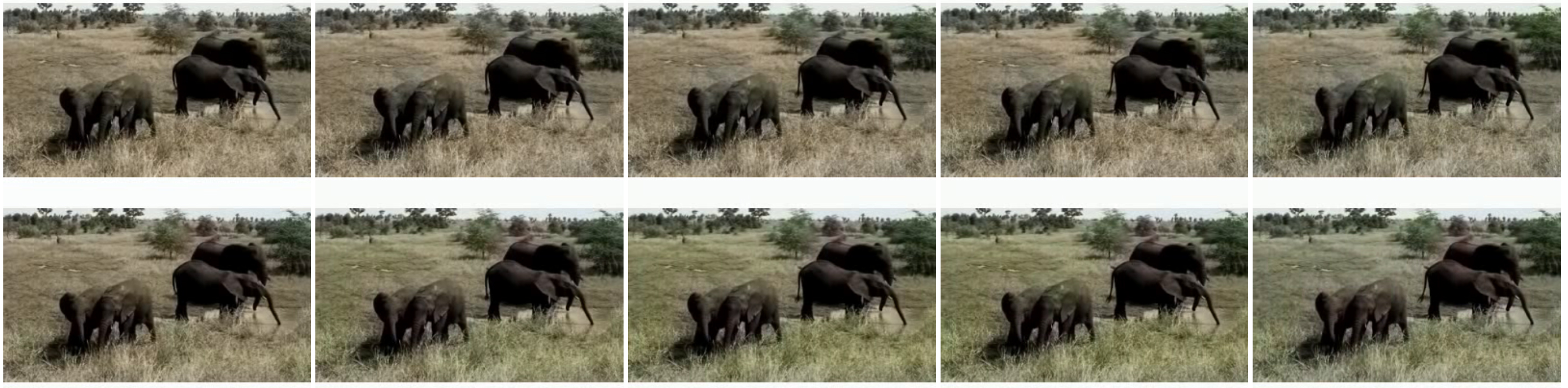}}
	\subcaptionbox{\label{fig:shooting}}
	  {\includegraphics[width=\textwidth]{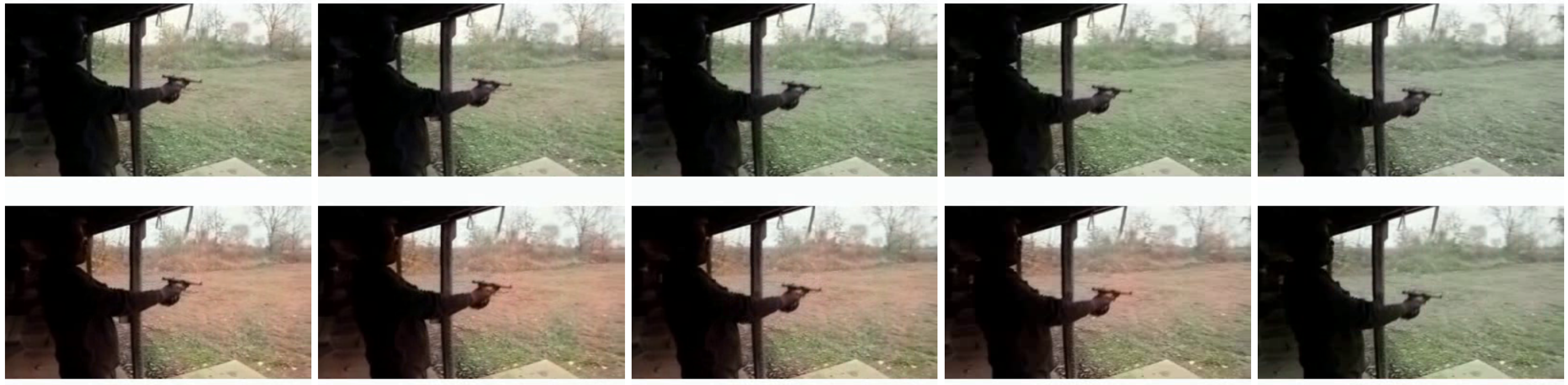}}
	\subcaptionbox{\label{fig:mobile-phone}}
	  {\includegraphics[width=\textwidth]{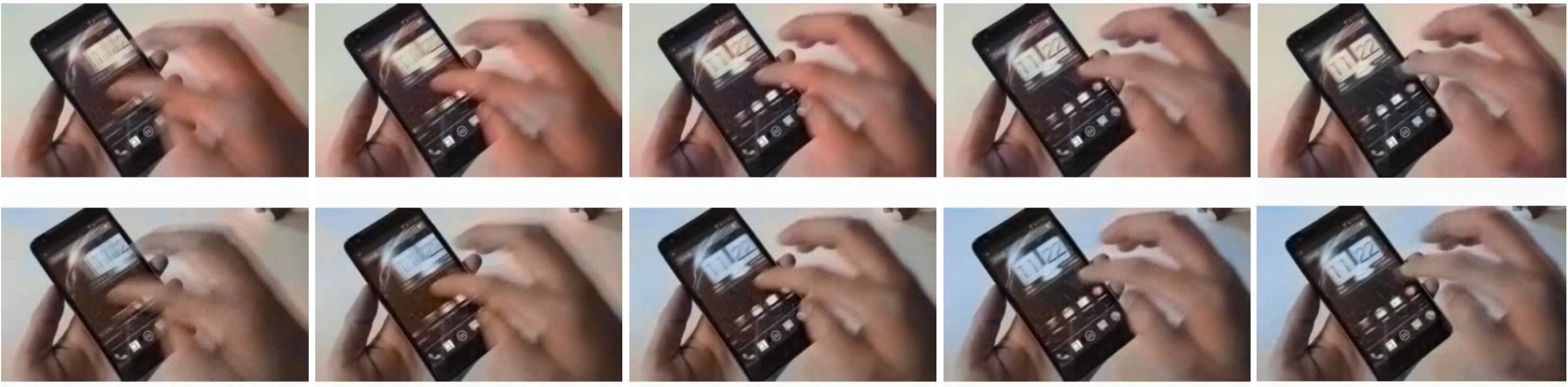}}
    
    \caption{In each sub-figure, the top and bottom rows show the video frame sequences colored by FlowChroma and the baseline model respectively. These show the superior global color palette maintenance throughout the scene by our model.}
\end{figure*}

An example of this is shown in Figure \ref{fig:elephants}, which depicts a herd of elephants strolling about. FlowChroma maintains the dry tone of the environment across the video while the baseline model shows colors changing between green and off-brown even for slight movements of elephants and their tails. Similarly, in \ref{fig:shooting}, FlowChroma again maintains the grass field in green while the baseline flickers from brown to green for the slight movements of the shooter and his gun. In \ref{fig:mobile-phone}, note how the baseline system bleeds color from the smartphone's clock into the background while our model does a better job of keeping the background uniform. 

\begin{figure*}[!h]
  \centering
  \subcaptionbox{\label{fig:baby-tub}}
    {\includegraphics[width=\textwidth]{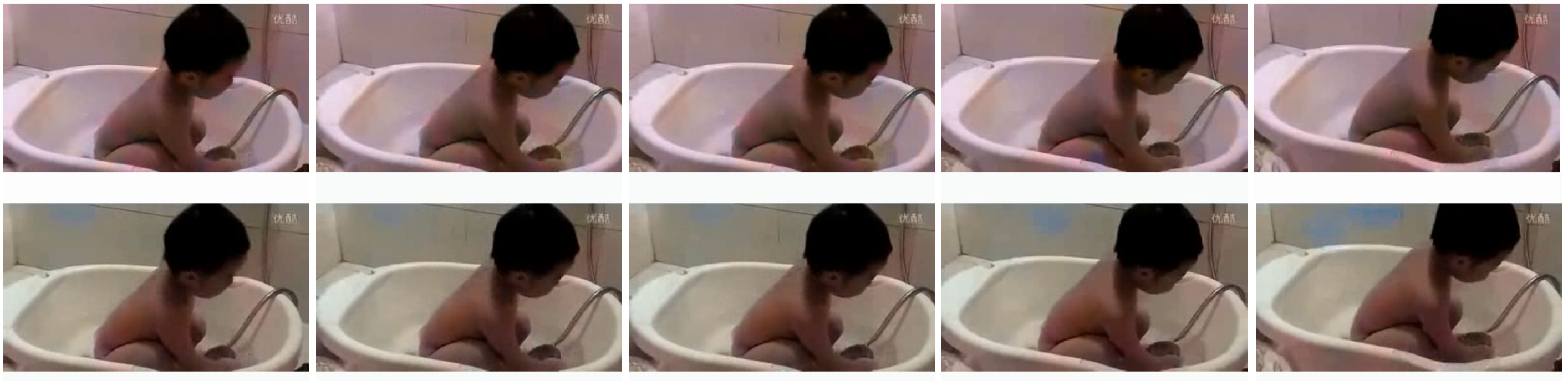}}
  \subcaptionbox{\label{fig:dk-playing}}
    {\includegraphics[width=\textwidth]{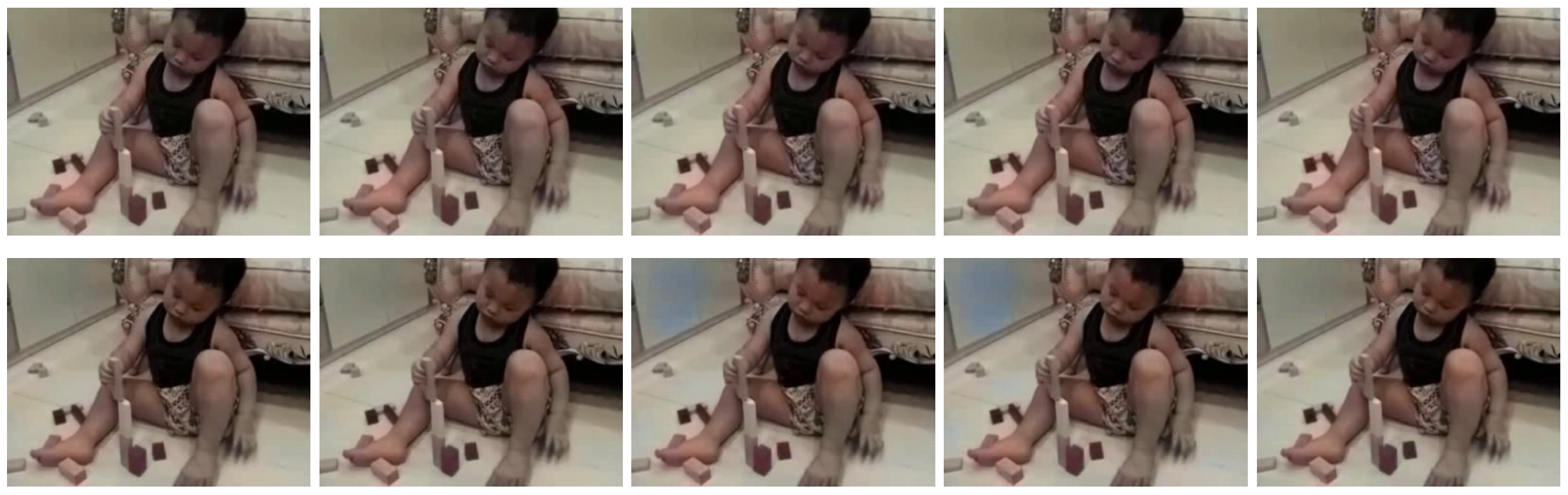}}
  \caption{In each sub-figure, the top and bottom row are from FlowChroma and the baseline model, respectively, showing how the local color uniformity is better maintained by FlowChroma. Note how the baseline model flickers with blue color patches as the camera angle changes in \subref{fig:baby-tub} and as the boy moves his hands in \subref{fig:dk-playing}.}
\end{figure*}

At a local scale, the LSTM affects how FlowChroma decides colors should be assigned even when it cannot fully identify the progression of a previously detected region. In Figure \ref{fig:baby-tub}, the background contains a wall that is uniformly colored throughout the frames by FlowChroma, while having blue patches in the baseline model's output. This is an example of the downsides of considering each frame as a separate colorization task, as done by image colorization models. Figure \ref{fig:dk-playing} contains an off-white board that is consistently colored by FlowChroma, whereas the baseline model again adds blue color patches. Blue color patches have appeared probably due to erroneously identifying those regions as sky or water in some frames.

Based on our observations, we can divide the factors affecting the consistency of colorization as temporal and non temporal. Non-temporal factors include 
\begin{enumerate}
\item extreme pixel values of input grayscale image e.g. extreme dark color of asphalt roads or extreme bright color of snow, 
\item the prevalence of a context in the training dataset. These factors affect both image colorization extensions to video colorization as well as FlowChroma. If the pixel values are extreme, such as in the case of snow or asphalt roads, both the baseline model and FlowChroma tend to leave them as extremes without assigning new colors. 
\end{enumerate}
Furthermore, when colorizing commonly encountered contexts, both the baseline and our model provided consistent appropriate colors because of the high level feature extractor; Inception-ResNet-v2 that is pre-trained on the ImageNet dataset, which contains images of commonly encountered context.

Temporal factors mainly relate to the movement of objects in a scenario, where the action frequency confuses the system's perception of the trajectory of the scene. This is applicable only to FlowChroma. When the movements in a video are smooth, our system identifies the objects and applies appropriate, temporally coherent coloring. When the movement in the scenario speeds up, the perceived flow of movement breaks and thus the colorization quality degrades fast, especially in terms of segmentation and appropriate coloring.

Lastly, we observe when the colorization of FlowChroma becomes inconsistent and also propose possible solutions for them. 
\begin{enumerate}
    \item The introduction of new objects into a scene changes its context, introducing momentary flickering before stabilizing again. Training the model further may alleviate this problem.
    \item When there is a high object frequency in a scene, the aptness of the colorization gets reduced. An example would be a surface with a complex pattern. A potential solution would be to train the system on more videos with high object frequency.
    \item The action frequency also adversely affects the system's performance. Normalizing the action speed is one possible solution. This could be done by increasing the number of frames containing the movement by predicting intermediate frames, as recently demonstrated by Nvidia \cite{DBLP:journals/corr/abs-1712-00080}, and then slowing down the video to achieve the desired speed. Another potential solution is to train the system with more time steps.
\end{enumerate}

\section{Conclusions}

Contemporary image colorization techniques are not directly applicable to video colorization as they treat each video frame as a separate colorization task, without maintaining temporal coherence between frames. We propose FlowChroma, a novel colorization framework with a recurrent neural network - LSTM - added to maintain temporal and contextual information between frames.

Inherent capability of LSTMs to learn how much each hidden cell should remember or forget while reading or generating a sequence stands as a justification for using LSTMs in FlowChroma rather than vanilla RNNs - this is the basis for their usage in video colorizations with scene changes.

We show that the LSTM maintains the image colorization quality of current methods intact while also successfully minimizing flickering between frames. It maintains the overall color palette of a scenario across subsequent frames at a global level, while coloring identified objects within a scene consistently at a local level. 

We observed some limitations in the use of recurrent architectures for video colorization, which may be common to other techniques as well. FlowChroma specifically generates inconsistent colorizations in the following scenarios; 
\begin{enumerate}
\item Sudden introduction of new objects into the scene
\item High object frequency or having high number of objects in a scene
\item High action frequency or fast movements in a scene.
\end{enumerate}

Finally, from these preliminary results, we have a promising research direction in maintaining temporal and contextual coherence in video colorization with LSTMs. As future work, we hope to quantitatively assess the performance of FlowChroma using a video colorization benchmark. We also plan to perform a visual Turing test of colorized videos from various frameworks.

%
%
%
%
%
%

\end{document}